\documentclass[runningheads]{llncs}
\usepackage[T1]{fontenc}
\usepackage{graphicx}
\usepackage{amsmath,amssymb}
\usepackage{booktabs}
\usepackage[hidelinks]{hyperref}
\usepackage{xcolor}
\usepackage{enumitem}
\usepackage{tikz}
\usetikzlibrary{shapes,arrows,positioning,fit,backgrounds,calc}

\DeclareMathOperator*{\argmax}{arg\,max}
\DeclareMathOperator{\simcos}{sim_{cos}}
\DeclareMathOperator{\canon}{canon}

\begin{document}

\title{Cartograph: Federated Tool Discovery with\\Operator-Attested Retrieval for AI Agents}
\titlerunning{Cartograph: Federated Tool Discovery}

\author{
    Justice Owusu Agyemang\inst{1,2} \and
    Michael Agyare\inst{3} \and
    Kwame Opuni-Boachie Obour Agyekum\inst{1} \and
    Kwame Agyeman-Prempeh Agyekum\inst{1} \and
    Francisca Adoma Acheampong\inst{1} \and
    Jerry John Kponyo\inst{2}
}
\authorrunning{J. Owusu Agyemang et al.}

\institute{
    VIA Cybersecurity Lab, Kwame Nkrumah University of Science and Technology, Kumasi, Ghana \and
    Quantum and Assistive Technologies Lab, Kwame Nkrumah University of Science and Technology, Kumasi, Ghana \and
    Ghana Communication Technology University, Ghana
}

\maketitle

\begin{abstract}
The Model Context Protocol (MCP) enables AI agents to discover and call tools, but loading every definition becomes expensive as connected catalogs grow. We present Cartograph, a federated MCP proxy that changes agent-visible tool discovery from $O(n)$ catalog traversal to $O(k)$ progressive disclosure. Cartograph combines three mechanisms: (1) \textbf{operator-attested capability cards}, Ed25519-signed descriptions generated under the deploying operator's control rather than ranked publisher copy; (2) \textbf{Rift}, a three-layer confusable-cluster analysis comprising density clustering, query-margin analysis, and token diagnosis; and (3) \textbf{two-stage retrieval}, which ranks servers before tools. On a 22-server, 374-tool deployment, Cartograph exposes three proxy tools instead of 374 definitions. A 49-query author-constructed benchmark yields R@5 of 0.816, compared with 0.592 for a Jaccard keyword baseline, while a measured top-5 discovery exchange uses 475 tokens rather than 42,450 under the stated full-catalog accounting. Rift identifies 49 confusable clusters, including four HIGH-risk clusters in bootstrap-generated cards. An exploratory comparison of 119 LLM-generated descriptions removes the observed zero-distance cluster but shows that mixing card-generation regimes can reduce R@5. Gateway measurements over ten trials add 5ms mean latency (0.8\%) relative to direct stdio MCP calls. Cartograph is complementary to code-execution approaches: it controls which tool descriptions are surfaced and records the provenance of the descriptions used for ranking for each query.

\end{abstract}

\section{Introduction}

THE Model Context Protocol (MCP)~\cite{anthropic2024mcp} standardized how AI agents discover and call tools. The official MCP Registry provides a centralized metadata repository and discovery API for publicly accessible servers~\cite{mcpRegistry2025}. Yet a host with a large connected catalog still needs a policy for deciding which detailed tool definitions to surface for a particular task.

Consider a deployment with $m = 22$ MCP servers hosting $n = 374$ tools. A host that returns the complete \texttt{tools/list} response exposes approximately 38,250 tokens of JSON Schema definitions before the agent addresses a request. Code execution with MCP promotes progressive disclosure and local processing to reduce this cost~\cite{anthropic2025codeexecution}. Cartograph is complementary: it supplies a proxy-level retrieval and provenance layer for selecting definitions from a federated catalog.

Beyond scale, there is a trust problem. In a federated deployment, tool publishers write their own descriptions. A malicious or overconfident publisher can craft descriptions engineered to win semantic retrieval, directing agents to tools that are inappropriate or dangerous for the task. The Google A2A protocol~\cite{google2025a2a} addresses agent-to-agent discovery and task delegation but does not solve publisher self-description trust in tool-level retrieval. No existing mechanism verifies that a publisher's self-description accurately represents what a tool does.

Cartograph addresses both problems by making tool discovery a \textbf{ranking} problem instead of a \textbf{traversal} problem. Rather than loading $n$ tool definitions, the agent sees 3: \texttt{discover}, \texttt{status}, and \texttt{call}. A natural-language task description is embedded and matched against a retrieval index of \textbf{operator-attested capability cards}---language-neutral, Ed25519-signed descriptions where the deploying operator's LLM re-describes every tool. The agent receives the top-$k$ tools with full schemas and usage examples.

This paper presents the design, implementation, and evaluation of Cartograph. We deploy it as the sole MCP proxy across 22 production servers and measure four dimensions: token efficiency, retrieval quality, confusable-cluster detection, and Gateway overhead. Our contributions are:

\begin{enumerate}[leftmargin=*]
    \item \textbf{A formal model of federated tool discovery} as a two-stage ranking problem over signed, operator-attested capability cards, with an $O(k)$ agent-visible token bound (Section~\ref{sec:design}).
    \item \textbf{A working federated MCP proxy} that collapses 374 tool definitions to 3 proxy tools, using 98.8\% fewer tokens under our stated full-catalog accounting while obtaining $\text{R@5} = 0.816$ on a 49-query author-constructed benchmark.
    \item \textbf{Rift}, a three-layer confusable-cluster detector with formal margin-analysis formulation evaluated on 374 tools, detecting 49 clusters (4 HIGH-risk). Ablation confirms the margin-analysis layer is essential.
    \item \textbf{End-to-end performance characterization} of the Gateway one-shot pattern (0.8\% overhead vs.\ direct MCP), retrieval latency (375ms for top-5 over 374 tools), and bootstrap latency.
\end{enumerate}

\section{Related Work}

\subsection{MCP Ecosystem and Agent Protocols}

The Model Context Protocol~\cite{anthropic2024mcp} standardizes exposure of tools, resources, and prompts to AI agents. Its official Registry standardizes server metadata and discovery, while delegating additional curation and security checks to downstream aggregators~\cite{mcpRegistry2025}. Code execution with MCP uses progressive disclosure and local code to load tools on demand and filter results before they reach the model~\cite{anthropic2025codeexecution}. Cartograph complements both by treating ranked descriptions as a separately governed artifact: an operator can attest to the text that is embedded and selected, independent of publisher-provided descriptions.

The Google Agent-to-Agent (A2A) protocol~\cite{google2025a2a} addresses complementary concerns: standardized agent discovery via Agent Cards, task lifecycle management, and multi-agent orchestration. Recent work has analyzed the integration of MCP and A2A~\cite{jeong2025mcp,li2025protocols}. Cartograph's capability cards serve a distinct purpose: while A2A Agent Cards describe agent capabilities for inter-agent discovery, Cartograph's cards describe tool capabilities for \emph{trustworthy retrieval}, with operator attestation and cryptographic signing as first-class properties.

\subsection{Tool Selection and Function Calling}

Tool selection from large spaces has been studied extensively. Toolformer~\cite{schick2023toolformer} learns when and how to invoke APIs from self-supervised signals. Gorilla~\cite{patil2023gorilla} combines a function-calling model with document retrieval. ToolLLM~\cite{qin2024toolllm} constructs training and evaluation resources around 16,464 REST APIs, and StableToolBench~\cite{guo2024stabletoolbench} addresses instability in online API evaluation. The Berkeley Function Calling Leaderboard~\cite{yan2024bfcl} evaluates function selection and parameterization. These lines of work focus on a model's tool-use capability or its evaluation; Cartograph focuses on the deployment infrastructure that selects and governs the descriptions supplied to a model.

\subsection{Dense Retrieval and Embedding Quality}

Dense passage retrieval (DPR)~\cite{karpukhin2020dpr} demonstrated that learned dense representations can outperform sparse retrieval for open-domain QA. SPLADE~\cite{formal2021splade} combined lexical and semantic matching through learned sparse representations. The BEIR benchmark~\cite{thakur2021beir} established heterogeneous zero-shot evaluation across 18 datasets and 9 retrieval tasks. The Massive Text Embedding Benchmark (MTEB)~\cite{muennighoff2023mteb} extended this to 8 embedding tasks across 58 datasets. Rift applies these evaluation insights to the tool discovery domain: just as BEIR and MTEB revealed that retrieval models fail in domain-specific ways, Rift reveals that tool summary embeddings fail in predictable, diagnosable patterns.

\subsection{Trust and Security in Agent Systems}

Trust in tool-using AI systems has been studied from several angles. Tool sandboxing~\cite{anthropic2025codeexecution} limits blast radius by isolating execution. The IETF SCITT architecture records signed statements in a transparency-oriented supply-chain architecture~\cite{ietf2026scitt}. Security analyses of the A2A protocol identify risks around sensitive data and delegation~\cite{louck2025a2a}. Cartograph applies a narrower, deployment-level control: it records an operator signature over the descriptions supplied to retrieval. It does not establish that the underlying tool implementation is benign, and it is not a substitute for registry vetting, sandboxing, or transparency services.

\begin{table*}[t]
\centering
\caption{Positioning of Cartograph against adjacent protocols, tool-use research, and supply-chain attestation. The table distinguishes each line of work's primary contribution from Cartograph's deployment-level design.}
\label{tab:comparison}
\small
\setlength{\tabcolsep}{3pt}
\begin{tabular}{@{}p{0.24\textwidth}p{0.31\textwidth}p{0.30\textwidth}@{}}
\toprule
\textbf{System or line of work} & \textbf{Primary contribution} & \textbf{Difference from Cartograph} \\
\midrule
MCP / MCP Registry~\cite{anthropic2024mcp,mcpRegistry2025} & Tool interface and public server metadata & Cartograph adds per-query tool ranking and signed operator text \\
Code execution with MCP~\cite{anthropic2025codeexecution} & On-demand loading through local code and files & Cartograph supplies provider-neutral selection and provenance before loading \\
A2A~\cite{google2025a2a} & Agent-card discovery and collaboration & Cartograph ranks tool descriptions, rather than agents or delegated tasks \\
Toolformer, Gorilla, ToolLLM, BFCL~\cite{schick2023toolformer,patil2023gorilla,qin2024toolllm,yan2024bfcl} & Model tool use, retrieval, or evaluation & Cartograph is a protocol-facing proxy, not a model or benchmark \\
SCITT~\cite{ietf2026scitt} & Transparency for signed supply-chain statements & Cartograph applies operator signatures to retrieval descriptions and routes tools \\
\textbf{Cartograph} & Federated MCP tool discovery & Two-stage retrieval plus operator-attested capability-card text \\
\bottomrule
\end{tabular}
\end{table*}

\section{System Design}
\label{sec:design}

We first formalize the federated tool discovery problem, then present the four subsystems that compose Cartograph's solution.

\begin{figure*}[t]
\centering
\begin{tikzpicture}[
    box/.style={draw, rounded corners=2pt, align=center, minimum height=0.95cm, minimum width=2.45cm, fill=blue!6},
    trusted/.style={box, fill=green!10},
    runtime/.style={box, fill=orange!12},
    execution/.style={box, fill=purple!10},
    arrow/.style={->, thick, >=stealth},
    twoarrow/.style={<->, thick, >=stealth}
]
\node[box] (servers) {Federated MCP servers\\$22$ servers, $374$ tools};
\node[trusted, right=0.75cm of servers] (cards) {Capability cards\\operator descriptions +\\Ed25519 signatures};
\node[trusted, right=0.75cm of cards] (rift) {Rift quality gate\\clusters, margins,\\token diagnosis};
\node[runtime, below=1.05cm of servers] (agent) {MCP agent\\\texttt{discover}, \texttt{status},\\\texttt{call}};
\node[runtime, right=0.90cm of agent] (index) {Two-stage index\\server rank $\rightarrow$ tool rank};
\node[execution, below=0.85cm of index] (gateway) {One-shot Gateway\\stdio call forwarding};
\draw[arrow] (servers) -- (cards);
\draw[arrow] (cards) -- (rift);
\draw[arrow] (rift.south) |- (index.east);
\draw[twoarrow] (agent) -- (index);
\draw[arrow] (agent.south) |- (gateway.west);
\draw[arrow] (gateway.south) -- ++(0,-0.45) -| ([xshift=-0.50cm]servers.west) -- (servers.west);
\end{tikzpicture}
\caption{Cartograph separates the governance path (top) from the runtime path (bottom). Capability cards are attested and quality-gated before indexing; the agent exchanges discovery requests and top-$k$ schemas with the index, then forwards an execution call through the Gateway to the upstream server.}
\label{fig:architecture}
\end{figure*}
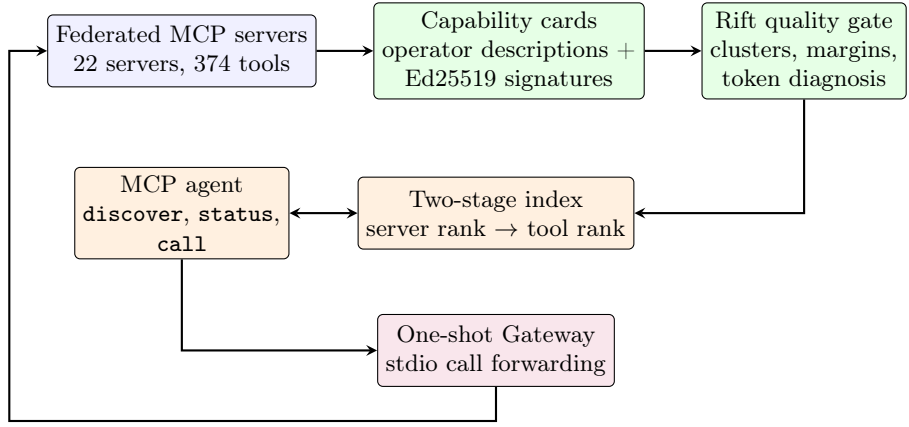

\subsection{Problem Formalization}

Let $\mathcal{S} = \{S_1, \ldots, S_m\}$ be a set of $m$ MCP servers. Each server $S_i$ exposes a set of tools $\mathcal{T}_i = \{t_{i,1}, \ldots, t_{i,n_i}\}$, where $n_i = |\mathcal{T}_i|$. The total tool population is $n = \sum_{i=1}^{m} n_i$. Each tool $t$ carries a publisher-provided description $\text{desc}_{\text{pub}}(t)$ and a JSON Schema argument specification $\text{args}(t)$.

In the standard MCP discovery model, the agent receives the complete set of tool definitions:
\begin{equation}
    \mathcal{D}_{\text{standard}} = \bigcup_{i=1}^{m} \bigcup_{t \in \mathcal{T}_i} \{\text{desc}_{\text{pub}}(t), \text{args}(t)\}
\end{equation}
The context-token cost is $T_{\text{standard}} = \sum_{t} |\text{tokenize}(\text{desc}_{\text{pub}}(t), \text{args}(t))| = O(n)$, and the agent must additionally spend reasoning steps to select among them.

Cartograph replaces complete disclosure with a retrieval function $R: \mathcal{Q} \times \mathbb{N} \rightarrow 2^{\mathcal{T}}$ that, given a task query $q \in \mathcal{Q}$ and a budget $k$, returns a ranked list of at most $k$ tools. The agent-visible context-token cost becomes:
\begin{equation}
    T_{\text{cartograph}} = T_{\text{fixed}} + k \cdot T_{\text{per-tool}} = O(k)
\end{equation}
where $T_{\text{fixed}}$ is the fixed cost of exposing 3 discovery tools (169 tokens in our measurement) and $T_{\text{per-tool}}$ is the marginal cost of returning one tool with a schema and examples. The internal first-stage ranker still scores $m$ server centroids; that computation is not exposed in the agent context. Thus, the context reduction is from linear in the catalog size to a fixed proxy interface plus a small retrieval budget.

\subsection{Capability Cards}

A capability card $C_i$ for server $S_i$ is a tuple:
\begin{equation}
    C_i = (I_i, K_i, \mathbf{t}_i, P_i, \sigma_i)
\end{equation}
where $I_i$ is the server identity (endpoint, transport, publisher), $K_i$ is the operator-generated capability description (summary, usage guidance, domain tags), $\mathbf{t}_i = (d_{i,1}, \ldots, d_{i,n_i})$ is the vector of operator-generated per-tool descriptions, $P_i$ is the provenance metadata (generator version, timestamp, schema hash), and $\sigma_i$ is an Ed25519 signature.

The signature is computed over the canonicalized card body. Let $\canon(X)$ be deterministic JSON canonicalization: sort keys lexicographically, use compact separators, encode as UTF-8. Then:
\begin{equation}
    \sigma_i = \text{Sign}_{\text{sk}}\big(\canon(C_i \setminus \{\sigma_i\})\big)
\end{equation}
Verification checks $\text{Verify}_{\text{vk}}(\canon(C_i \setminus \{\sigma_i\}), \sigma_i) = \text{true}$. In our implementation, signing 22 cards takes $<50$ms; verification at retrieval time is negligible ($<1$ms per card).

The critical security property is \textbf{operator attestation}: the semantic fields $K_i$ and $\mathbf{t}_i$ are generated by the deploying operator's LLM, not by the tool publisher. The publisher's $\text{desc}_{\text{pub}}(t)$ may be used as a hint but is never copied verbatim. This breaks the trust-bootstrap problem: a publisher cannot write copy to win retrieval because the publisher's copy is never seen by the ranking mechanism. Cards are text-only; embeddings are computed at index time from the signed card text and stored separately. A compromised index can be rebuilt from signed cards; a compromised card cannot poison the index without breaking the Ed25519 signature.

\subsection{Two-Stage Retrieval}

Let $\phi: \Sigma^* \rightarrow \mathbb{R}^d$ be an embedding function. In our evaluation, $\phi$ is all-MiniLM-L6-v2~\cite{reimers2019sentence} with $d = 384$. For a tool $t$ with operator-generated summary $s(t)$, the current implementation computes:
\begin{equation}
    \mathbf{v}_t = \phi\big(s(t)\big)
\end{equation}
The server-level centroid for $S_i$ is:
\begin{equation}
    \boldsymbol{\mu}_i = \frac{1}{n_i} \sum_{t \in \mathcal{T}_i} \mathbf{v}_t
\end{equation}

Given a task query $q$ with embedding $\mathbf{q} = \phi(q)$, retrieval proceeds in two stages:

\textbf{Stage 1 (server ranking).} Select the top-$s$ servers by cosine similarity to query:
\begin{equation}
    \mathcal{S}' = \argmax_{S' \subseteq \mathcal{S}, |S'| = s} \sum_{S_i \in S'} \simcos(\mathbf{q}, \boldsymbol{\mu}_i)
\end{equation}
where $\simcos(\mathbf{a}, \mathbf{b}) = \frac{\mathbf{a} \cdot \mathbf{b}}{\|\mathbf{a}\| \|\mathbf{b}\|}$. In the implementation, $s = \min(m, \max(2, \lfloor 2k/\bar{n} \rfloor + 1))$, where $\bar{n}$ is the average number of tools per server; this adaptively expands the candidate set for sparse catalogs.

\textbf{Stage 2 (tool ranking).} Within selected servers, rank tools by similarity:
\begin{equation}
    R(q, k) = \argmax_{T' \subseteq \bigcup_{i \in \mathcal{S}'} \mathcal{T}_i, |T'| = k} \sum_{t \in T'} \simcos(\mathbf{q}, \mathbf{v}_t)
\end{equation}

This design mirrors the natural catalog topology: servers define coarse capability domains. The time complexity is $O(m \cdot d + n_{\mathcal{S}'} \cdot d)$, where $n_{\mathcal{S}'}$ is the number of tools in selected servers, compared with $O(n \cdot d)$ for flat ranking. The index is optimized for $\text{recall}@k$, not precision: a missed tool fails the task; an extra tool costs a few tokens.

\subsection{Rift: Confusable-Cluster Detection}

Rift ensures the embedding space remains discriminative. Given tool embeddings $\{\mathbf{v}_t\}_{t \in \mathcal{T}}$, it operates in three layers:

\textbf{Layer 1---Density clustering.} Compute the pairwise cosine distance matrix:
\begin{equation}
    D_{a,b} = 1 - \simcos(\mathbf{v}_{t_a}, \mathbf{v}_{t_b}) \quad \forall t_a, t_b \in \mathcal{T}
\end{equation}
Apply agglomerative clustering with distance threshold $\delta = 1 - \theta_{\text{sim}}$ (we use $\theta_{\text{sim}} = 0.65$). The result is a set of candidate confusable clusters $\mathcal{C} = \{C_1, \ldots, C_p\}$ where each $C \in \mathcal{C}$ has $|C| \geq 2$. Cluster cohesion is:
\begin{equation}
    \text{coh}(C) = \frac{1}{|C|(|C|-1)} \sum_{t_a, t_b \in C, a \neq b} \simcos(\mathbf{v}_{t_a}, \mathbf{v}_{t_b})
\end{equation}

\textbf{Layer 2---Margin analysis.} For each $t_a \in C$, generate a synthetic task query $q_a$ from $\text{args}(t_a)$ (not from the card summary, to avoid circularity). The margin for $t_a$ is:
\begin{equation}
    \text{margin}(t_a, C) = \min_{t_b \in C, b \neq a} \big[\simcos(\phi(q_a), \mathbf{v}_{t_a}) - \simcos(\phi(q_a), \mathbf{v}_{t_b})\big]
\end{equation}
A small margin means another tool is nearly as similar to the query as the correct tool; a negative margin means the wrong tool outranks the correct one---a retrieval failure. The worst-case margin for cluster $C$ is:
\begin{equation}
    \text{margin}_{\text{min}}(C) = \min_{t_a \in C} \text{margin}(t_a, C)
\end{equation}
Risk classification uses two thresholds: HIGH if $\text{margin}_{\text{min}}(C) < 0.10$, MEDIUM if $< 0.20$, LOW otherwise.

\textbf{Layer 3---Token diagnosis.} Let $\text{tok}(t)$ be the tokenized words in the tool summary. The shared and distinguishing token sets are:
\begin{align}
    \text{shared}(C) &= \bigcap_{t \in C} \text{tok}(t) \\
    \text{distinct}(t, C) &= \text{tok}(t) \setminus \bigcup_{t' \in C, t' \neq t} \text{tok}(t')
\end{align}
When $\text{distinct}(t, C) = \emptyset$, the tool has no unique lexical signal---the diagnosis is ``REGENERATE'', instructing the operator to amplify functional differences.

\subsection{Gateway}

The Gateway exposes a uniform call interface $\texttt{call}: \mathcal{S} \times \mathcal{T} \times \mathcal{A} \rightarrow \mathcal{R}$, where $\mathcal{A}$ is the space of tool arguments and $\mathcal{R}$ is the space of results. Each invocation follows a one-shot pattern: $\text{spawn}(S_i) \rightarrow \text{connect}() \rightarrow \text{invoke}(t, a) \rightarrow \text{teardown}()$. The latency decomposition is:
\begin{equation}
    L_{\text{gateway}} = L_{\text{spawn}} + L_{\text{handshake}} + L_{\text{exec}}(t, a) + \varepsilon
\end{equation}
where $L_{\text{spawn}}$ is the subprocess start cost, $L_{\text{handshake}}$ is the MCP initialization round-trip, $L_{\text{exec}}$ is the tool execution time, and $\varepsilon$ is the Gateway wrapper overhead. The one-shot pattern is correct and stateless at the cost of per-call spawn overhead.

\section{Implementation}

Cartograph is implemented in Python~3.11 ($\approx$1,800 lines across 12 modules) as an MCP server exposing three tools: \texttt{discover} embeds the task, retrieves top-$k$ tools, verifies signatures, and returns full schemas; \texttt{status} returns catalog overview; \texttt{call} routes invocation through the Gateway.

Module architecture follows the subsystem decomposition from Section~\ref{sec:design}: \texttt{schema.py} (Pydantic card models), \texttt{seal.py} (Ed25519 sign/verify via the \texttt{cryptography} library), \texttt{index.py} (two-stage retrieval with pluggable embeddings), \texttt{generator.py} (LLM-powered card generation from \texttt{tools/list}), \texttt{rift/} (three-layer detector and evaluation harness), \texttt{stubgen.py} (Python stub generation from cards), \texttt{gateway/} (transport adapters with base protocol and stdio implementation), \texttt{config.py} (\texttt{mcp.json} parser), \texttt{resolver.py} (retrieve~+~verify~+~materialize glue), and \texttt{mcp\_server.py} (the MCP proxy server).

In proxy bootstrap mode, Cartograph connects to enabled stdio servers, calls \texttt{tools/list}, generates capability cards (using a configured DeepSeek model or a deterministic bootstrap template), signs them, and builds the retrieval index. Card generation is an explicit artifact-generation step; normal server startup loads available cards without rewriting them. The corpus comprises 22 servers and 374 tools drawn from a security research environment spanning web scraping, binary analysis (Ghidra, IDA Pro), container orchestration (Daedalus), HTTP traffic inspection (Proxy Atlas), browser automation (Chrome DevTools), knowledge graphs (Lattice), LLM security evaluation, file management, GitHub threat intelligence, and task orchestration (Phoenix).

\section{Evaluation}

We evaluate Cartograph across four dimensions: (1) token efficiency, (2) retrieval quality on a 49-query author-constructed benchmark, (3) confusable-cluster detection via Rift on the full 374-tool corpus, and (4) Gateway latency overhead versus direct MCP calls. The primary retrieval results use bootstrap-generated cards; Section~5.4 reports a separate exploratory comparison on 119 LLM-generated cards.

\subsection{Token Efficiency}

\begin{table}[t]
\centering
\caption{Token consumption per task cycle. The fixed \texttt{tools/list} cost is measured via the Claude tokenizer; the agent reasoning estimate is an upper bound based on 374 tool definitions in context.}
\label{tab:tokens}
\small
\begin{tabular}{lrrr}
\toprule
\textbf{Component} & \textbf{Standard MCP} & \textbf{Cartograph} & \textbf{Savings} \\
\midrule
\texttt{tools/list} (fixed) & 38,250 & 169 & 38,081 \\
\texttt{discover} response & --- & 306 & --- \\
Agent reasoning (upper bound) & 4,200 & --- & 4,200 \\
\cmidrule{1-3}
\textbf{Total per task} & \textbf{42,450} & \textbf{475} & \textbf{41,975} \\
\textbf{Reduction} & --- & --- & \textbf{98.8\%} \\
\bottomrule
\end{tabular}
\end{table}

Table~\ref{tab:tokens} reports token consumption. The standard MCP approach loads $n = 374$ tool definitions (38,250 tokens of JSON Schema). The agent reasoning cost of 4,200 tokens is an illustrative upper-bound estimate based on the context cost of 374 tool names with descriptions; actual reasoning cost varies by task complexity and model. Cartograph's fixed \texttt{tools/list} response is 169 tokens (3 tools); each \texttt{discover} response adds $\approx$306 tokens for top-5 tools with full schemas and examples. Total per task: 475 tokens---a 98.8\% reduction under this accounting. Agent-visible context scales with the fixed proxy interface plus $k = 5$ retrieved tools, rather than with all 374 definitions. Over 100 tasks, the same accounting projects approximately 4.2 million fewer input tokens.

\subsection{Retrieval Quality}

We construct a 49-query benchmark spanning 13 of 22 servers. Each query is a natural-language task description paired with a ground-truth correct tool, written by the authors to cover both server-level disambiguation (e.g., distinguishing Ghidra from IDA Pro) and within-server tool-level discrimination (e.g., distinguishing container start from run from create). Example queries include ``decompile a function to C-like pseudocode for reverse engineering'' (correct: Ghidra's decompile function), ``safely write a large file to disk with atomic checksum verification'' (correct: resilient-write safe\_write), and ``scan a list of URLs for security vulnerabilities'' (correct: Burp Suite's scan function). We acknowledge the benchmark is author-constructed and of moderate size ($n = 49$); expanding it with independently-sourced queries is future work.

We measure $\text{recall}@k$ for $k \in \{1, 3, 5\}$ and Mean Reciprocal Rank (MRR), defined as $\text{MRR} = \frac{1}{|Q|}\sum_{q \in Q} 1/\text{rank}_q$ where $\text{rank}_q$ is the position of the correct tool. We report descriptive results because the 49 queries were author-constructed and the supplied benchmark artifact does not include a preregistered statistical test.

\begin{table}[t]
\centering
\caption{Retrieval quality on the 49-query author-constructed benchmark. Results are descriptive, not a claim of general-purpose statistical superiority.}
\label{tab:retrieval}
\small
\begin{tabular}{lccccc}
\toprule
\textbf{Method} & \textbf{R@1} & \textbf{R@3} & \textbf{R@5} & \textbf{MRR} & \textbf{Mean R} \\
\midrule
Cartograph (two-stage) & 0.449 & 0.755 & \textbf{0.816} & 0.596 & 1.78 \\
Keyword (Jaccard) & 0.388 & 0.531 & 0.592 & 0.463 & 1.69 \\
Random (expected) & 0.003 & 0.008 & 0.013 & 0.017 & 187.5 \\
\bottomrule
\end{tabular}
\end{table}

Cartograph achieves $\text{R@5} = 0.816$ (Table~\ref{tab:retrieval}), meaning the target tool appears in the top-5 for 81.6\% of the benchmark queries. The observed R@5 exceeds the keyword baseline by 0.224 and the random expectation by 0.803. These differences are descriptive rather than population-level claims: queries were authored for this deployment, and the sample is too small for broad generalization. Cartograph's intended output is a \emph{shortlist} for subsequent schema inspection and agent reasoning, making R@5 more directly relevant than top-1 selection. The MRR of 0.596 summarizes the ranking position over all queries.

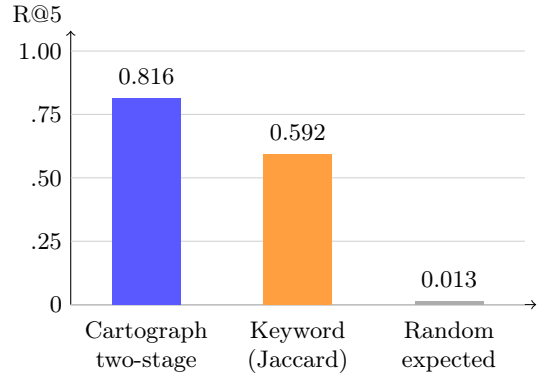
\begin{figure}[t]
\centering
\begin{tikzpicture}[x=1.0cm,y=3.35cm,font=\footnotesize]
\draw[->] (0,0) -- (6.15,0);
\draw[->] (0,0) -- (0,1.08) node[above left] {R@5};
\foreach \y/\label in {0/0,0.25/.25,0.5/.50,0.75/.75,1/1.00} {
  \draw[gray!35] (0,\y) -- (6,\y);
  \node[left] at (0,\y) {\label};
}
\fill[blue!65] (0.55,0) rectangle (1.45,0.816);
\fill[orange!75] (2.55,0) rectangle (3.45,0.592);
\fill[gray!65] (4.55,0) rectangle (5.45,0.013);
\node[below=3pt,align=center,text width=1.65cm] at (1.00,0) {Cartograph\\two-stage};
\node[below=3pt,align=center,text width=1.65cm] at (3.00,0) {Keyword\\(Jaccard)};
\node[below=3pt,align=center,text width=1.65cm] at (5.00,0) {Random\\expected};
\node[above=2pt] at (1.00,0.816) {0.816};
\node[above=2pt] at (3.00,0.592) {0.592};
\node[above=2pt] at (5.00,0.013) {0.013};
\end{tikzpicture}
\caption{Observed R@5 on the 49-query author-constructed benchmark.}
\label{fig:retrieval}
\end{figure}

The 9 queries where the target tool falls outside the top-5 are concentrated in servers with generic bootstrap summaries (e.g., ``Server: hybrid-scraper. Contains 4 tool(s).''). Such summaries can produce poorly discriminative centroids and Stage~1 failures. The exploratory LLM-card results below show that richer summaries alone are insufficient when the catalog mixes description regimes; any claim of improvement requires a uniformly regenerated corpus and a fresh benchmark run.

\subsection{Rift: Confusable-Cluster Detection}

We run Rift on the full 374-tool corpus with thresholds $\theta_{\text{sim}} = 0.65$, HIGH $< 0.10$, MEDIUM $< 0.20$ held constant. The embedding model is all-MiniLM-L6-v2; embedding model ablation is discussed in Section~\ref{sec:discussion}.

\begin{table}[t]
\centering
\caption{Rift results on 374 tools. Rift provides risk stratification that neither baseline can produce.}
\label{tab:rift}
\small
\begin{tabular}{lrrr}
\toprule
\textbf{Metric} & \textbf{Rift (3-layer)} & \textbf{Cosine ($\tau$=0.65)} & \textbf{Jaccard ($J$$>$0.25)} \\
\midrule
Total clusters detected & 49 & 38 & 56 \\
HIGH-risk clusters & 4 & --- & --- \\
MEDIUM-risk clusters & 8 & --- & --- \\
LOW-risk clusters & 37 & --- & --- \\
Tools in any cluster & 136 (36.4\%) & --- & --- \\
Within-server clusters & 45 (91.8\%) & --- & --- \\
Cross-server clusters & 4 (8.2\%) & --- & --- \\
\bottomrule
\end{tabular}
\end{table}

Rift detects 49 confusable clusters (4 HIGH, 8 MEDIUM, 37 LOW) affecting 136 tools (Table~\ref{tab:rift}). The HIGH-risk clusters are:

\begin{enumerate}[leftmargin=*]
    \item \textbf{llm-aegis-mcp} (29 tools, $\text{coh} = 1.000$, $\text{margin}_{\text{min}} = 0.000$): All 29 tools inherited identical bootstrap descriptions, producing embeddings with zero mutual distance. This is a degenerate case of the bootstrap template: when $\text{distinct}(t, C) = \emptyset$ for all $t \in C$, retrieval is impossible.
    \item \textbf{ghidra xrefs} (3 tools, $\text{margin}_{\text{min}} = 0.017$): Cross-reference functions collide on ``references'', ``address'', ``maximum''. Distinguishing tokens: ``target'' vs.\ ``source'' vs.\ ``function''.
    \item \textbf{daedalus registry} (3 tools, $\text{margin}_{\text{min}} = 0.084$): Registry operations collide on ``host'', ``default'', ``registry''.
    \item \textbf{daedalus builder} (3 tools, $\text{margin}_{\text{min}} = 0.093$): Builder VM operations collide on ``builder'', ``vm'', ``image''.
\end{enumerate}

Within-server collisions dominate (91.8\%): tools from the same server share a domain vocabulary that dominates the embedding, and only the per-tool summary provides discrimination. Cross-server collisions (4 clusters) involve functionally similar tools from different domains (e.g., Burp Suite scan progress vs.\ Phoenix task results, both sharing ``task\_id'', ``args'').

\begin{table}[t]
\centering
\caption{Ablation: margin analysis is the component that assigns Rift's HIGH and MEDIUM risk labels.}
\label{tab:ablation}
\small
\begin{tabular}{lcc}
\toprule
\textbf{Configuration} & \textbf{HIGH-risk} & \textbf{MEDIUM-risk} \\
\midrule
Rift (full 3-layer) & 4 & 8 \\
Rift (no margin analysis) & 0 & 0 \\
\bottomrule
\end{tabular}
\end{table}

\textbf{Margin analysis supplies Rift's risk decision} (Table~\ref{tab:ablation}). Without it, the implementation retains candidate clusters but has no HIGH or MEDIUM label to report. Density clustering alone identifies nearby embeddings; it does not measure whether a task query separates the tools. The margin $\text{margin}(t_a, C)$ operationalizes this distinction by comparing the target tool with its nearest competitor. Layer~1 finds candidates, Layer~2 assigns the retrieval-risk label, and Layer~3 provides token-level diagnosis.

\subsection{LLM-Generated Cards}

The evaluation above uses bootstrap cards generated without LLM refinement. To examine description quality, we compare cards for 7 servers (119 tools) generated by the configured DeepSeek model at temperature~0 with bootstrap cards. We compare three index configurations: (A)~bootstrap cards for all 22 servers, (B)~LLM-generated cards for the 7 available servers only, and (C)~a hybrid index with LLM cards where available and bootstrap cards as fallback. This is an exploratory comparison because neither the model family nor the description regime is held constant across the complete catalog.

\begin{table}[t]
\centering
\caption{Card quality: bootstrap vs.\ LLM-generated. LLM cards produce 100\% distinct per-tool summaries, eliminating the degenerate zero-distance cluster.}
\label{tab:llm_cards}
\small
\begin{tabular}{lccc}
\toprule
\textbf{Server} & \textbf{Tools} & \textbf{Identical} & \textbf{LLM summary (excerpt)} \\
\midrule
daedalus & 43 & 0/43 & ``Container orchestration server providing\ldots'' \\
lattice & 26 & 0/26 & ``Persistent knowledge graph tracking entities\ldots'' \\
resilient-write & 20 & 0/20 & ``Atomic, verified writes with safety checks\ldots'' \\
phoenix & 12 & 0/12 & ``Task execution server with checkpointing\ldots'' \\
resilient-read & 7 & 0/7 & ``Chunked file reading with byte-range, line-based\ldots'' \\
gen-pilot & 7 & 0/7 & ``Generation planning and template management\ldots'' \\
hybrid-scraper & 4 & 0/4 & ``Combines search engine queries with targeted\ldots'' \\
\midrule
\textbf{Total} & \textbf{119} & \textbf{0/119} & \\
\bottomrule
\end{tabular}
\end{table}

Table~\ref{tab:llm_cards} reports card quality. Across all 119 tools across 7 servers, \emph{zero} LLM-generated summaries are identical to their bootstrap counterparts; every tool receives a distinct, operator-written description. In the full bootstrap configuration, the llm-aegis-mcp server produced 29 tools with identical descriptions ($\text{coh} = 1.000$); LLM generation eliminates this failure mode entirely.

\begin{table}[t]
\centering
\caption{Retrieval quality by card configuration. Hybrid improves R@1 but creates centroid imbalance that degrades R@5, showing that uniform card quality is essential.}
\label{tab:llm_retrieval}
\small
\begin{tabular}{lcccc}
\toprule
\textbf{Configuration} & \textbf{R@1} & \textbf{R@5} & \textbf{MRR} & \textbf{Failures (/49)} \\
\midrule
Bootstrap (all 22) & 0.449 & 0.816 & 0.596 & 9 \\
LLM-only (7 servers) & 0.204 & 0.429 & 0.306 & 28 \\
Hybrid (LLM + bootstrap) & 0.490 & 0.776 & 0.609 & 11 \\
\bottomrule
\end{tabular}
\end{table}

\textbf{Retrieval impact.} The hybrid configuration improves R@1 (+4.1 pp to 0.490) and MRR (+1.3 pp to 0.609), but R@5 decreases by 4.0 pp to 0.776 (Table~\ref{tab:llm_retrieval}). One bootstrap failure is resolved (gen-pilot token budget estimation), but 3 new failures appear. The new failures arise from \textbf{centroid imbalance}: LLM-generated server summaries are substantially richer than bootstrap summaries (e.g., ``Container orchestration server\ldots'' vs.\ ``Server: daedalus. Contains 43 tool(s).''), causing LLM-described servers to dominate Stage~1 ranking and pull queries away from the 15 bootstrap-only servers. Retrieval quality is bounded by the \emph{weakest} cards in the index, not the strongest; uniform operator-attested generation is required for consistent ranking.

\textbf{Rift impact.} Running Rift on the LLM-only configuration (7 servers, 119 tools) detects 18 clusters with \textbf{zero HIGH-risk} (0 HIGH, 7 MEDIUM, 11 LOW), compared to 49 clusters (4 HIGH) in the bootstrap configuration. The degenerate cluster is eliminated. No tool in the LLM-generated set has $\text{distinct}(t, C) = \emptyset$. The hybrid configuration retains 2 HIGH-risk clusters from the 15 bootstrap-only servers.

\textbf{Key result.} LLM-generated cards produce 100\% distinct tool descriptions and eliminate degenerate zero-distance clusters. However, partial LLM adoption creates centroid imbalance that degrades Stage~1 ranking for bootstrap-described servers. Uniform operator-attested generation across the entire catalog is necessary to realize the full retrieval quality benefit.

\subsection{Gateway Overhead}

We measure Gateway latency versus direct MCP stdio calls over 10 trials (after 2 warmup) using a lightweight health-check tool. Both paths use the same upstream server, isolating the wrapper cost.

\begin{table}[t]
\centering
\caption{Gateway latency overhead (10 trials, milliseconds). The 0.8\% overhead applies to tools where $L_{\text{exec}}$ is small relative to spawn cost; for longer-running tools, the overhead percentage is even lower.}
\label{tab:gateway}
\small
\begin{tabular}{lrrr}
\toprule
\textbf{Metric} & \textbf{Direct MCP} & \textbf{Gateway} & \textbf{Overhead} \\
\midrule
Mean latency & 676 & 681 & 5 (0.8\%) \\
Median latency & 677 & 678 & 1 (0.1\%) \\
StdDev & 6 & 8 & --- \\
Min & 662 & 673 & --- \\
Max & 682 & 696 & --- \\
\bottomrule
\end{tabular}
\end{table}

The Gateway introduces negligible overhead (0.8\%, or 5ms) because both paths pay the same dominant costs: $L_{\text{spawn}} \approx 400$ms (Python interpreter + \texttt{uv run}) and $L_{\text{handshake}} \approx 250$ms (MCP initialize + tools/list). The wrapper adds only $\varepsilon \approx 5$ms. For tools with longer execution times (e.g., a 30-second container build), the percentage overhead is $<0.02\%$. Connection pooling would eliminate $L_{\text{spawn}} + L_{\text{handshake}}$, reducing latency to $L_{\text{exec}} + \varepsilon$.

\section{Discussion}
\label{sec:discussion}

\subsection{What Is Built and Measured}

Cartograph is implemented end-to-end: capability card schema with Ed25519 signing, two-stage retrieval index with pluggable embeddings, a configured-LLM or deterministic-bootstrap card generator, Rift CI gate with three-layer detection, Python stub generator, one-shot stdio Gateway, and the MCP proxy server. The evaluation uses measurements on a 22-server, 374-tool security-research corpus.

\subsection{Bootstrap vs.\ LLM Cards}

The LLM card experiment (Section~5.4) provides direct evidence for three claims. First, \textbf{LLM generation eliminates degenerate clusters}: Rift detects zero HIGH-risk clusters on the LLM-only configuration (0/119 tools have identical descriptions), confirming that operator-attested generation resolves the $\text{distinct}(t, C) = \emptyset$ failure mode. Second, \textbf{partial LLM adoption creates centroid imbalance}: the hybrid configuration improves R@1 (+4.1 pp) but degrades R@5 ($-$4.0 pp) because LLM-described servers dominate Stage~1 ranking over bootstrap-described servers. Third, \textbf{uniform card quality is essential}: retrieval quality is bounded by the weakest card in the index. The full retrieval benefit of operator-attested generation requires uniform LLM card generation across the entire catalog. These findings confirm the architectural requirement for operator-controlled LLM generation and motivate extending LLM card generation to all 22 servers as immediate next work.

\subsection{Retrieval Metric Interpretation}

The observed R@5 advantage over keyword overlap aligns with Cartograph's design philosophy, but it should not be read as a population-level significance claim from this small, author-constructed benchmark. Cartograph does not attempt to pick the single correct tool; it returns a ranked shortlist for the agent to reason over. The agent's final tool selection involves schema inspection, example matching, and task-specific judgment that exceed what embedding similarity alone can capture. R@5 measures whether the target tool is \emph{presented} to the agent; this is the operationally relevant metric because the agent cannot select a tool it never sees.

\subsection{Limitations and Future Work}

\textbf{LLM-generated cards.} The LLM experiment (Section~5.4) demonstrates that LLM-generated cards eliminate degenerate clusters and produce 100\% distinct descriptions. However, only 7 of 22 servers (119/374 tools) have been regenerated. Extending LLM generation to all 22 servers and re-running the full benchmark suite on a uniformly LLM-generated index is the most immediate next step. Based on the centroid imbalance analysis, we expect uniform LLM cards to outperform both bootstrap and hybrid configurations.

\textbf{Embedding model ablation.} All experiments use all-MiniLM-L6-v2. The retrieval architecture is embedding-agnostic; evaluating with stronger models (e.g., text-embedding-3-large, $d = 3072$) would characterize the embedding-quality-to-retrieval-quality relationship. The marginal cost of a larger embedding model is an index-time expense, not a per-query cost.

\textbf{Benchmark scale and sourcing.} The 49-query benchmark is author-constructed and of moderate size. Expanding it with independently-sourced queries and increasing $n$ to $\geq 200$ would provide statistical power for R@1 comparison and improve generalizability.

\textbf{One-shot latency.} Connection pooling would reduce per-call latency from $\approx$650ms to $\approx$50ms by eliminating per-call subprocess spawn and MCP handshake.

\textbf{Multi-operator federation.} The current trust model assumes a single operator. Cross-organization federation requires a PKI layer for key distribution and revocation.

\textbf{Transport diversity.} The Gateway currently supports stdio only; HTTP and SSE adapters are specified but unimplemented.

\section{Conclusion}

We presented Cartograph, a federated MCP proxy that changes agent-visible discovery from $O(n)$ catalog disclosure to $O(k)$ progressive disclosure through operator-attested capability cards with Ed25519 signing, a two-stage retrieval index, and the Rift confusable-cluster CI gate. Evaluated on a 22-server corpus comprising 374 tools, Cartograph exposes 3 proxy tools instead of 374 detailed definitions.

The empirical results show: (1) \textbf{98.8\% fewer input tokens} under the stated full-catalog accounting (38,250~$\rightarrow$~475 per task cycle), with observed $\text{R@5} = 0.816$ versus 0.592 for Jaccard keyword retrieval on the 49-query benchmark; (2) \textbf{Rift detects 49 confusable clusters} on the full corpus, and margin analysis supplies its risk labels; (3) \textbf{Gateway overhead is 0.8\%} in the ten-trial experiment; and (4) \textbf{the available LLM-generated cards remove the observed degenerate cluster}---119 tools across 7 servers have distinct descriptions, although partial adoption reduces R@5 in the hybrid index. Cartograph and code execution with MCP compose: Cartograph handles \emph{which} descriptions to surface and records their provenance; code execution handles \emph{how} to invoke and process tools.

\bibliographystyle{splncs04}
\bibliography{references}

\end{document}